\documentclass{article}
\usepackage{amsmath,epsfig}
\usepackage{booktabs}
\usepackage{amsthm}
\usepackage{amsmath,amssymb,amsfonts}
\usepackage{algorithmic}
\usepackage{algorithm}
\newtheorem{theorem}{Theorem}

\newtheorem{remark}{Remark}

\usepackage[preprint]{spconfa4}

\copyrightnotice{\textbf{978-1-6654-3864-3/21/\$31.00~\copyright 2021 IEEE}}

\let\OLDthebibliography\thebibliography
\renewcommand\thebibliography[1]{
  \OLDthebibliography{#1}
  \setlength{\parskip}{0pt}
  \setlength{\itemsep}{0pt plus 0.3ex}
}

\begin{document}\sloppy

\def\x{{\mathbf x}}
\def\L{{\cal L}}

\title{Cooperative Learning for Noisy Supervision}
%
\name{Hao Wu$^{1}$\thanks{This work is supported by the National Key Research and Development Program of China (No. 2019YFB1804304), SHEITC (No. 2018-RGZN-02046), 111 plan (No. BP0719010),  and STCSM (No. 18DZ2270700), and State Key Laboratory of UHD Video and Audio Production and Presentation.}, Jiangchao Yao$^{1}$, Ya Zhang$^{1,2,*}$
\thanks{$^{*}$Corresponding author}
, Yanfeng Wang$^{1,2}$}

\address{$^{1}$Cooperative Medianet Innovation Center, Shanghai Jiao Tong University, China\\
$^{2}$Shanghai Artificial Intelligence Laboratory, China\\
\{howiethepeanut, sunarker, ya\_zhang, wangyanfeng\}@sjtu.edu.cn
}

\maketitle

\begin{abstract}
Learning with noisy labels has gained the enormous interest in the robust deep learning area. Recent studies have empirically disclosed that utilizing dual networks can enhance the performance of single network but without theoretic proof. In this paper, we propose Cooperative Learning (CooL) framework for noisy supervision that analytically explains the effects of leveraging dual or multiple networks. Specifically, the simple but efficient combination in CooL yields a more reliable risk minimization for unseen clean data. A range of experiments have been conducted 
on several benchmarks with both synthetic and real-world settings. Extensive results indicate that CooL outperforms several state-of-the-art methods.

\end{abstract}
\begin{keywords}
deep learning, noisy supervision, cooperative learning
\end{keywords}
\section{Introduction}
\label{sec:intro}

Large-scale supervised training datasets have significantly driven the success of deep neural networks 
(DNNs) in computer vision
~\cite{krizhevsky2012imagenet,szegedy2015going,he2016deep}.
However, accurate annotations provided by the human experts are usually expensive to collect in many real-world applications.
Therefore, the alternative ways such as crowdsourcing~\cite{raykar2010learning} and web query~\cite{fergus2010learning} 
are explored to reduce the cost. For example, there are a plethora of images tagged by users on open platforms which can be exploited easily and inexhaustibly. The negative impact is these approaches also inevitably introduce label noise to the dataset since the user annotations are not very reliable. Considering DNNs are capable of fitting extremely noisy labels~\cite{zhang2016understanding}, it is important to make the training robust to such label noise.

Recent studies 
can be roughly categorized into two classes in terms of \emph{the classifier count}, i.e., single-network structure and dual-network structure.
For the former, it usually refers to robust surrogate losses, noise transition and self-paced learning. For instance, Bootstrap~\cite{reed2014training} applies the perceptional consistency to the cross-entropy loss to mitigate the influence of label noise. Forward~\cite{patrini2017making} build a transition matrix on top of the classifier to absorb the noise. Self-paced MentorNet~\cite{jiang2018mentornet} selects small loss samples as clean instances and learn only from these instances. For the latter, it introduces twin classifiers to be the teacher of each other by which boosts the classification performance of the single network. Decouple~\cite{malach2017decoupling} utilizes the prediction disagreement of twin networks to select more informative samples as supervision. CLC~\cite{wu2019collaborative} leverages the entropy criterion to collaboratively correct the labels. Co-distillation~\cite{anil2018large,song2018collaborative} distills the knowledge of the one to supervise 
the other one and vice versa. Co-teaching~\cite{han2018co} leverages two networks to select small loss instances for cross update.



Although the current dual-network structure empirically shows improvement over the single version, it lacks of theoretical analysis and guarantee that it can always work. Besides, a natural question is whether introducing more learners can further benefit the learning with noisy supervision. 
To explore these limits and give a more general scope, we propose a Cooperative Learning (CooL) paradigm that multiple classifiers work cooperatively for noisy supervision. Specifically, we firstly demonstrate the dual-network structure yields lower risk than that associated with the single network in some suitable cooperation. 
Then, we give a sufficient condition for the case of more learners, where the risk is negatively correlated to the number of classifiers. 
Generally, even although the classifiers are imperfect with noisy supervision, a lower risk can be achieved when the more disagreement is introduced. Finally, based on these analysis, a cooperative learning framework is introduced that the cooperation supervision is utilized to improve the performace. The main contribution can be summarized into the following three points. We demonstrate in the presence of noisy supervision, the linear combination of predictions from multiple networks yields a more reliable supervision than predictions from either single classifier in some conditions. A Cooperative Learning framework is introduced, where multiple different imperfect classifiers produce supervision cooperatively to iteratively boost the performance. We empirically verify the proposed method on CIFAR-10, CIFAR-100 with synthetic noise and three large-scale real-world datasets namely Clothing1M, Food-101N and WebVision. Comprehensive experiments show
that CooL outperforms several state-of-the-art methods.

\section{The Proposed Method}
\subsection{Preliminaries}
Given a noisy dataset $\mathcal{D}=\{(x_n, y_n)\}^N_{n=1}$, where $N$ is the sample number, $x_n$ denotes an image instance and $y_n\in \{0,1\}^c$ is the corresponding noisy label, we target to produce more reliable supervision for the classifiers to learn in the presence of label noise. Assume we use $f_i(i=1,2,\dots)$ to represent a classifier with index $i$ and $p_{i}$ indicates the prediction of $f_i$. We explore to achieve this goal via the cooperation of multiple classifiers.

\subsection{Dual-Network Cooperative Learning}
As empirically indicated in several works~\cite{anil2018large,song2018collaborative,han2018co,malach2017decoupling}, the dual-network structure is easy to acquire a robust classification performance when learning with noisy supervision. To understand the law of this phenomenon, we deduce the theoretical analysis in 
light
of risk minimization. 
We term this methodology as dual-network Cooperative Learning (CooL) to ease the explanation and unify the notion of this work. 

Suppose there are two classifiers $f_1$ and $f_2$ and we use the combination of the predictions $p_1$ and $p_2$ respectively output from $f_1$ and $f_2$ as the new cooperation supervision:

\begin{align}\label{eq:pseudo_label}
    \hat{p}^\lambda = \lambda {p_1} + (1-\lambda){p_2}.
\end{align}
$\lambda$ is the cooperation parameter to balance between the predictions from the two classifiers.
To measure the reliability of a certain supervision $\Tilde{y}$ with respect to the ground-truth label $y^*$, 
we define the 
noisy supervision risk on the training set $r_{\Tilde{y}} = E_{\mathcal{D}}[\|\Tilde{y}-y^*\|^2].$ 
\footnote{
We take inspiration from Li et al.~\cite{li2017learning} while our scenarios are different. We propose to measure risks on the training set with no clean labels.
 }
In the following, we will show that with a suitable choice of $\lambda$. leveraging the cooperation supervision in Eq.~\eqref{eq:pseudo_label} yields lower risk than the individual risk of the either model.

\begin{theorem}\label{th:th1}
There always exists a $\lambda$ that makes 
the risk $r_{\hat{p}^\lambda}$ of the dual-network cooperation lower than\footnote{We exclude the situation where one classifier strictly dominates the other otherwise there is no need for cooperation since $\lambda$ will be set to 0 or 1 and the cooperation risk will be equal to the lower one.} individual risks of two non-identical networks that do not always produce the incorrect predictions at the same time, i.e., 
\begin{align}
    \exists \lambda,~r_{\hat{p}^\lambda} < \min\{r_{p_1}, r_{p_2}\},\nonumber
\end{align}
\end{theorem}

\begin{proof}
First, the risks of the individual predictions from $f_1$ and $f_2$ are quantified respectively as the following terms,
\begin{align}
\begin{split}
    & r_{p_1} = E_{\mathcal{D}}[\|p_1-y^*\|^2],
    r_{p_2} = E_{\mathcal{D}}[\|p_2-y^*\|^2].\nonumber \\
\end{split}
\end{align}
Then, the cooperation risk can be decomposed as follows, \begin{align}
\begin{split}
     r_{\hat{p}^\lambda_i}
     & = E_{\mathcal{D}}[\|\hat{p}^\lambda-y^*\|^2] 
     = E_{\mathcal{D}}[\|\lambda {p_1} + (1-\lambda){p_2}-y^*\|^2] \\
     & = \lambda^2r_{p_1} + 2\lambda(1-\lambda)r_{p_1p_2} + (1-\lambda)^2r_{p_2}.\nonumber
\end{split}
\end{align}
where
$r_{p_1p_2} \triangleq E_{\mathcal{D}_t}[({p_1}-y^*)^T({p_2}-y^*)]$. For two non-identical classifiers $f_1$ and $f_2$ that do not always produce the incorrect predictions at the same time, as $p_1$, $p_2$ are label distributions and $y^*$ is the one-hot label, we will have
\begin{align} \label{eq:r_cross}
0 \leq r_{p_1p_2} < \min\{r_{p_1},r_{p_2}\}.
\end{align}
By setting 
$\nabla_\lambda r_{\hat{p}^\lambda_i}=0$, 
we obtain
\begin{align}
\begin{split}
    \min_{\lambda}r_{\hat{y}^\lambda} 
    &= r_{p_1}-\frac{(r_{p_1}-r_{p_1p_2})^2}{r_{p_1}+r_{p_2}-2r_{p_1p_2}} \\
    &= r_{p_2}-\frac{(r_{p_2}-r_{p_1p_2})^2}{r_{p_1}+r_{p_2}-2r_{p_1p_2}} \\
    &< \min\{r_{p_1}, r_{p_2}\}.
\end{split}\label{eq:dif}
\end{align}
which concludes the proof of Theorem~\ref{th:th1}.
\end{proof}

\begin{remark}\label{re:re1}
From Theorem~\ref{th:th1}, we know the suitable dual-network cooperation can achieve a lower risk than the individual network. Furthermore, it can be found the minimum risk obtained in Eq.~\eqref{eq:dif} is positively correlated to $r_{p_1p_2}$. This term reflects divergences between the two classifiers on the samples where both of the classifiers make incorrect predictions. The optimal situation is that two classifiers never make mistakes at the same, then we have $r_{p_1p_2} = 0$. 
Intuitively, both $p_1$ and $p_2$ are deviated from the true label $y^*$. However, these deviations are towards random directions in the presence of stochastic label noise, the proposed cooperation supervision can be closer to the true label. 
\end{remark}

\subsubsection{Connection and Difference.}
Here, we rethink two representative dual-network methods namely Co-distillation and Co-teaching through the lens of Cooperative Learning. 

Co-distillation represents a branch of studies which leverage the model predictions to rectify the noisy labels. It is the case where $p_1$ in Eq.~\eqref{eq:pseudo_label} is substituted to the noisy label $y$.
Correspondingly setting $\lambda=\frac{r_{p_2}}{r_{y}+r_{p_2}}$, the optimal risk for $f_1$ can be simplified as $\frac{r_{y}r_{p_2}}{r_{y}+r_{p_2}}$, which is lower than the risk $r_{p_2}$. This explains why Co-distillation shows improvement. Nevertheless, it also points out one defect that $r_y$ is fixed and cannot be improved along with the decrease of $r_{p_2}$.

Co-teaching represents a line of studies which select samples with a certain criterion for training. Thus, we can adjust the risk by modifying the correlated dataset (removing the unreliable samples). According to~\cite{han2018co}, the supervision of $f_1$ is a candidate set $\mathcal{D}_{c_2}$ selected by $f_2$ and vice versa. The risk for $f_1$ is then denoted as $E_{\mathcal{D}_{c_2}}[\|y-y^*\|^2]$. Generally, it is a lower risk than that on $\mathcal{D}$ with the help of the small loss trick. If we linearly combine the supervision like CooL, the cooperation risk is also a linear combination with respect to $\lambda$. And the minimum will be obtained at the boundaries, i.e., the smaller one in $r_{f_1}$ and $r_{f_2}$. In this case, a more reasonable way is to choose one of the two candidate sets, which has a lower risk. In Co-teaching, they utilize both sets for cross update which may impair the performance.

\subsection{Generalized Cooperative Learning}

In this section, we aim to generalize our dual-network CooL to a multi-network variant, which is able to achieve a even lower risk.
Given $n$ non-identical classifiers, we denote the new cooperation supervision as $\hat{p}^{\boldsymbol{\lambda}}=\boldsymbol{\lambda}\boldsymbol{p}$,
where $\boldsymbol{\lambda}$ is a $n$-dimension row vector with summation equal to 1 and $\boldsymbol{p}$ is a stack of $p_{1},\dots, p_{n}$ in rows. We define $\boldsymbol{R} = (r_{ij})_{n\times n}$
where $r_{ij} = E_{\mathcal{D}}[({p_i}-y^*)^T({p_j}-y^*)]$. The corresponding diagonal elements are the individual risks associated with the predictions of the $n$ classifiers respectively. In the following, we analyze the cooperation risk for multiple classifiers.

\begin{theorem}\label{th:th2} Given the cooperation supervision $\hat{p}^{\boldsymbol{\lambda}}=\boldsymbol{\lambda}\boldsymbol{p}$, the associated risk is $r_{\hat{p}^{\boldsymbol{\lambda}}}=\boldsymbol{\lambda}\boldsymbol{R}\boldsymbol{\lambda}^T$. 
An invertible $\boldsymbol{R}$ yields,
\begin{align}\label{eq:minr}
    \min_{\boldsymbol{\lambda}}r_{\hat{p}^{\boldsymbol{\lambda}}} = \frac{1}{\sum_{i,j=1}^{n}[\boldsymbol{R}^{-1}]_{i,j}}.
\end{align}
If all non-identical classifiers are independently trained in the same settings, so that the following conditions satisfy
\begin{align}\label{eq:condition}
\begin{split}
    r_{ii} &= r_{diag}, \forall i = 1,\dots,n\\
    r_{ij} &= r_{off}, \forall i\neq j
\end{split}
\end{align}
Then, the minimum cooperation risk in Eq.~\eqref{eq:minr} will be 
\begin{align}\label{eq:cooln}
    \min_{\boldsymbol{\lambda}}r_{\hat{p}^{\boldsymbol{\lambda}}} 
    = \frac{1}{n}(r_{diag}-r_{off})+r_{off} < r_{diag},
\end{align}
which is lower than the individual risks of all classifiers.
\end{theorem}

\begin{proof}
The risk associated with the new supervision is
\begin{align}
\begin{split}
     r_{\hat{p}^{\boldsymbol{\lambda}}}
     & = E_{\mathcal{D}}[\|\boldsymbol{\lambda}\boldsymbol{p}-y^*\|^2]
     = E_{\mathcal{D}}[\|\sum_{i=1}^n\lambda_i ({p_i}-y^*)\|^2]\\
     & = \boldsymbol{\lambda}\boldsymbol{R}\boldsymbol{\lambda}^T, \textit{~~s.t.}, (1,\dots,1)\boldsymbol{\lambda}^T = 1.\nonumber\\ 
\end{split}
\end{align}
Leveraging the Lagrange multipliers, we now minimize,
\begin{align}
    \min f(\boldsymbol{\lambda}, \boldsymbol{\mu}) = \boldsymbol{\lambda}\boldsymbol{R}\boldsymbol{\lambda}^T - \boldsymbol{\mu}((1,\dots,1)\boldsymbol{\lambda}^T - 1)\nonumber
\end{align}
By setting $\frac{\partial f(\boldsymbol{\lambda}, \boldsymbol{\mu})}{\partial \boldsymbol{\lambda}}
=0$ and $\frac{\partial f(\boldsymbol{\lambda}, \boldsymbol{\mu})}{\partial \boldsymbol{\mu}}
 =0$
, we obtain, 
\begin{align}
    \boldsymbol{\lambda}_0 &= \frac{\boldsymbol{R}^{-1}(1,\dots,1)^T}{\sum_{i,j=1}^{n}[\boldsymbol{R}^{-1}]_{i,j}},\nonumber
    \boldsymbol{\mu}_0 = \frac{2}{\sum_{i,j=1}^{n}[\boldsymbol{R}^{-1}]_{i,j}}\nonumber.
\end{align}
Thus, the minimum risk associated with $r_{\hat{p}^{\boldsymbol{\lambda}}}$ is
\begin{align}
\begin{split}
    \min_{\boldsymbol{\lambda}}r_{\hat{p}^{\boldsymbol{\lambda}}} 
    &= f(\boldsymbol{\lambda}_0, \boldsymbol{\mu}_0)
    = \frac{1}{\sum_{i,j=1}^{n}[\boldsymbol{R}^{-1}]_{i,j}}.  \nonumber
\end{split}
\end{align}
To further analyze the characteristics of above equation, if we have Eq.~\eqref{eq:condition} satisfied
Eq.~\eqref{eq:minr} will be deduced as
\begin{align}
\begin{split}
    \min_{\boldsymbol{\lambda}}r_{\hat{p}^{\boldsymbol{\lambda}}} 
    &= \frac{1}{n}(r_{diag}-r_{off})+r_{off}
    < r_{diag}.\nonumber
\end{split}
\end{align}
which concludes the proof of Theorem~\ref{th:th2}.
\end{proof}

\begin{remark}\label{re:re2}
From Theorem~\ref{th:th2}, we can see that the first term on the RHS of Eq.~\eqref{eq:cooln} indicates by leveraging more classifiers (increasing n), we can monotonically obtain a lower risk. In this case, $\boldsymbol{R}$ being diagonally dominant yields a necessary and sufficient condition where the risk is inversely proportional to the number of the classifiers. Besides, similar to the claim in Remark~\ref{re:re1}, the off-diagonal element $r_{off}$ in $\boldsymbol{R}$ characterizes the divergence of two networks. If two classifies are complementary with each other, they will work better cooperatively even though they are imperfect. 
\end{remark}

\subsection{The Cooperative Learning Framework}\label{sec:framework}
The theoretical analysis in previous section tells us that the cooperation of multiple classifiers can lower the supervision risk and the lower bound is determintered by the divergence between the classifiers. Based on this, we introduce a new Cooperative Learning (CooL) 
framework where the proposed cooperation supervision namely the combination of the predictions from the multiple classifiers is adopted to re-train the individual networks. 
As claimed in Remark~\ref{re:re1} and~\ref{re:re2}, the prerequisite of a better performance on noisy datasets via cooperation, is to generate diverse classifiers. We thus make the classifiers learn from different sources of information to construct pattern bias. Specifically, we pre-train the $n$ classifiers respectively on $\mathcal{D}_1,\dots, \mathcal{D}_n$, the different $n$ partitions of $D$. 
Note this pre-training style relies on the assumption that the subset is still sufficient enough to learn a classifier exhibiting the same risk on the whole dataset. Thus we are not able to infinitely add classifiers to lower the risk due to limited data.

After obtaining multiple different pre-trained classifiers, we can utilize the combination of their predictions to train better classifiers. Instead of training another student network with such supervision like~\cite{li2017learning}, we iteratively train the classifiers with the objective function as follows,
\begin{align}
    L(f_i) &= l_{\mathcal{D}}(\hat{p}^\lambda, f_i) + \alpha l_{\mathcal{D}_i}(y,f_i) + \beta h_{\mathcal{D}_i}(f_i).\label{eq:Lf}
\end{align}
In the first term of Eq.~\eqref{eq:Lf}, the network $f_i$ is supervised by the cooperation supervision $\hat{p}^\lambda$, which is the key module of this paper. 
The second part is an auxiliary term that supervises $f_i$ with the original labels in the early phase but will be gradually canceled out as the model is capable of memorizing the noisy labels. The third term is the entropy of the model predictions which prevents the output of the network $f_i$ from degenerating to the uniform distribution. As for the hyperparameters $\alpha$ and $\beta$, we empirically assign small weights like~\cite{song2018collaborative}. The complete training process is summarized in Algorithm~\ref{alg:coolalgorithm}.

 \begin{algorithm}[tb]
 \scriptsize
\caption{The CooL Algorithm}\label{alg:coolalgorithm}
\begin{algorithmic}[1] 
\REQUIRE A noisy set $\mathcal{D}$, multiple networks $f_{1},\dots, f_{n}$, $\lambda$, $\alpha$, $\beta$. 
\STATE Randomly partition $\mathcal{D}$ into $\mathcal{D}_1,\dots, \mathcal{D}_n$. 
\STATE Directly pre-train $f_i$ on $\mathcal{D}_i(i=1,2,\dots)$ respectively.
\FOR{epoch $i=\text{StartEpoch}$ to $\text{MaxEpoch}$ }
\FOR{batch $j=1$ to $\frac{|\mathcal{D}|}{\text{BatchSize}}$ }
\FOR{$k=1$ to $n$}
\STATE Update $f_k$ by optimizing $L(f_k)$ in Eq.~\eqref{eq:Lf}
\ENDFOR
\ENDFOR
\ENDFOR
\end{algorithmic}
\end{algorithm}

\textbf{Complexity Analysis }
The time complexity of CooL is not a big issue since we can distribute the computation into individual classifiers parallelly. Assume $M$ is the mini-batch size and $\Lambda$ is the parameter size, then in each mini-batch update, the time complexity for each classifier is $\mathcal{O}(M\Lambda)$. However, for the space complexity, it might be a bottleneck as the storage cost is linearly related to the number of the classifier, i.e., $\mathcal{O}(nM\Lambda)$. Thus, when implementing multiple-network CooL in practices, we have to consider the resource limit. 

\section{Experiments}

\subsection{Datasets and Baselines}
\begin{table*}
\tiny
\centering
\begin{tabular}{lrrrrrr|rrrrrr} 
\toprule
 & \multicolumn{6}{c}{\textbf{CIFAR-10}} & \multicolumn{6}{c}{\textbf{CIFAR-100}}\\
\midrule
 & \multicolumn{2}{c}{\textbf{pairwise}} & \multicolumn{2}{c}{\textbf{asymmetric}} & \multicolumn{2}{c}{\textbf{symmetric}} & \multicolumn{2}{c}{\textbf{pairwise}} & \multicolumn{2}{c}{\textbf{asymmetric}} & \multicolumn{2}{c}{\textbf{symmetric}}\\
 \midrule
\textbf{setting} & 1 & 2 & 3 & 4 & 5 & 6 & 7 & 8 & 9 & 10 & 11 & 12\\
\midrule
\textbf{noise ratio ($r$)} & 0.2 & 0.45 & 0.2 & 0.45 & 0.2 & 0.5 & 0.2 & 0.45 & 0.2 & 0.45 & 0.2 & 0.5\\
\midrule
Standard & 76.54 & 49.73 & 82.86 & 69.52 & 76.71 & 49.91 & 50.47 & 32.51 & 50.79 & 31.47 & 48.63 & 25.44\\
Forward & 78.06 & 58.69 & 83.29 & 69.59 & 77.56 & 51.70 & 52.67 & 30.14 & 54.99 & 28.41 & 47.00 & 25.94\\
Bootstrap & 76.19 & 49.75 & 82.90 & 70.23 & 76.66 & 48.83 & 51.01 & 31.17 & 50.35 & 32.42 & 47.87 & 24.74\\
MentorNet & 80.32 & 58.67 & 83.62 & 70.28 & 80.73 & 68.70 & 50.89 & 32.26 & 50.27 & 32.35 & 52.12 & 37.96\\
\midrule
Decouple & 77.30 & 49.23 & 83.53 & 70.24 & 78.03 & 48.97 & 51.28 & 31.49 & 50.50 & 31.72 & 46.19 & 23.29\\
Co-distillation & 81.36 & 51.87 & 85.28 & 72.15 & 81.49 & 56.76 & 56.40 & 34.93 & 55.27 & 35.27 & 54.36 & 32.41\\
Co-teaching & 83.24 & 72.74 & 85.12 & 76.02 & 82.09 & 74.06 & 54.74 & 34.08 & 52.81 & 34.77 & 53.84 &  41.34\\
Bagging & 81.35 & 57.22 & 82.58 & 72.12 & 79.08 & 67.90 & 45.99 & 26.21 & 47.24 & 26.46 & 43.79 &  23.98\\
\midrule
CooL & \textbf{89.48} & \textbf{88.62} & \textbf{89.73} & \textbf{82.95} & \textbf{87.88} & \textbf{81.76} & \textbf{63.64} & \textbf{48.38} & \textbf{63.65} & \textbf{49.24} & \textbf{60.36} & \textbf{45.93}\\
\bottomrule
\end{tabular}
\caption{Average test accuracy (\%) on CIFAR-10 \& CIFAR-100 over the last ten epochs.}\label{tab:cifar}
\end{table*}

To demonstrate the effectiveness of CooL, we experiment with CIFAR10 and CIFAR100~\cite{krizhevsky2009learning} with pairwise noise~\cite{han2018co}, asymmetric noise~\cite{patrini2017making}, symmetric noise~\cite{van2015learning}
 and Clothing1M~\cite{xiao2015learning}, Food-101N~\cite{lee2018cleannet}, WebVision~\cite{li2017webvision} for real-world noise.
We compare CooL with the following two categories of noisy-supervised learning methods.
\textbf{Single-network Methods:} 
\emph{Standard}, which directly trains a vanilla classifier on noisy datasets; 
\emph{Forward}~\cite{patrini2017making}, which uses a noise transition matrix for the forward loss correction; 
\emph{LCCN}~\cite{yao2019safeguarded}, which dynamically adjust the transition matrix to safeguard the learning process. We only directly report the result of LCCN on WebVision to save huge labor to reproduce as we adopt the same settings;
\emph{Bootstrap}~\cite{reed2014training}, which linearly combines the model predictions and original labels; 
\emph{MentorNet}~\cite{jiang2018mentornet}. We deploy self-paced MentorNet namely a single model determines small loss samples as useful information and learn with these samples;
\textbf{Dual-network Methods:} 
\emph{Decouple}~\cite{malach2017decoupling}, which updates the parameters when the two models disagree; \emph{Co-distillation}~\cite{anil2018large,song2018collaborative}, which is the dual network version of Bootstrap; \emph{Co-teaching}~\cite{han2018co}, which is the dual network version of self-paced MentorNet; \emph{Bagging}~\cite{breiman1996bagging} which takes a vote of multiple classifiers pre-trained on random data partitions. Specifically we use two classifiers. 

\subsection{Implementation}

For CIFAR-10 and CIFAR-100, we follow the same implementation in~\cite{han2018co} 
For real-world datasets, a 50-layer ResNet architecture~\cite{he2016deep} pre-trained on ImageNet is adopted as the classifier. 
The images are resized to 256 with respect to shorter sides
and then randomly cropped 
to 224$\times$224
with random flip, brightness, contrast and saturation. 
For Clothing1M and Food-101N, 
the batch size is set to 64 and we run 20 epochs on Clothing1M and 40 epochs on Food-101N. For WebVision, we align the learning settings with Yao et al.~\cite{yao2019safeguarded} to save the labor in reproducing the results.

For all experiments, we use the same architecture for two networks when implementing Decouple, Co-teaching, Co-distillation and CooL as done in~\cite{malach2017decoupling,song2018collaborative,han2018co}. 
To be fair, all methods use pre-training
as warming-up following~\cite{patrini2017making,song2018collaborative,han2018co}. Specifically the models are trained as Standard for 4 epochs on Clothing1M and 10 epochs on all other datasets. For dual network methods, two branches are pre-trained separately.
For Forward, 
we use the normalized ground-truth confusion matrix provided in~\cite{xiao2015learning}
on Clothing1M. For MentorNet and Co-teaching, the noise ratio $r$ is provided as side information as required in~\cite{han2018co} for the pre-defined curriculum. 
For CooL, we set $\lambda=0.5$ since two networks have the same architecture and are trained in the same manner. We empirically set $\alpha=0.05$ on CIFAR-10 and Clothing1M while we set $\alpha=0.1$ on datasets containing much more categories. For the CIFAR datasets, we linearly decrease $\alpha$ since deep models easily fit the small-scale datasets and we fix $\beta=0.05$.

\subsection{Results on CIFAR10 and CIFAR-100}


Table~\ref{tab:cifar} summarizes the average test accuracy of dual network CooL and all baselines on CIFAR-10 and CIFAR-100 over the last ten epochs. We can see from the results that dual-network methods such as Co-distillation and Co-teaching generally perform better than their single version namely Bootstrap and MentorNet. Bagging works well on CIFAR-10 but fails on CIFAR-100. The reason is that CIFAR-100 contains more classes thus the data after random partition is insufficient to train a good individual learner. However, adopting similar style of pre-training, CooL manages to achieve the best performance in all of the noise settings. This indicates that CooL can effectively boost two imperfect individual learners. Pointedly for low-level pairwise noise 
, CooL outperforms the best baseline by 6.24\% on CIFAR-10 and 7.24\% on CIFAR-100. When $r$ raises to 0.45, all baselines degenerate hard while CooL shows great robustness dealing with high-level noise. According to the results, CooL outperforms the best baseline by 15.88\% on CIFAR-10 even impressively surpassing the results of all baselines under low-level noise. For asymmetric noise, CooL achieves the best performances in all settings and manages to outperform the best baseline by 14.47\% when $r=0.45$ on CIFAR-100. Symmetric noise is the
 hardest noise pattern
  as the overall test accuracies are low. However, CooL manages to outperform all the baselines.

\subsubsection{Counteracting the Memorization Effects}

\begin{figure} 
\centering
\includegraphics[width=0.5\textwidth]{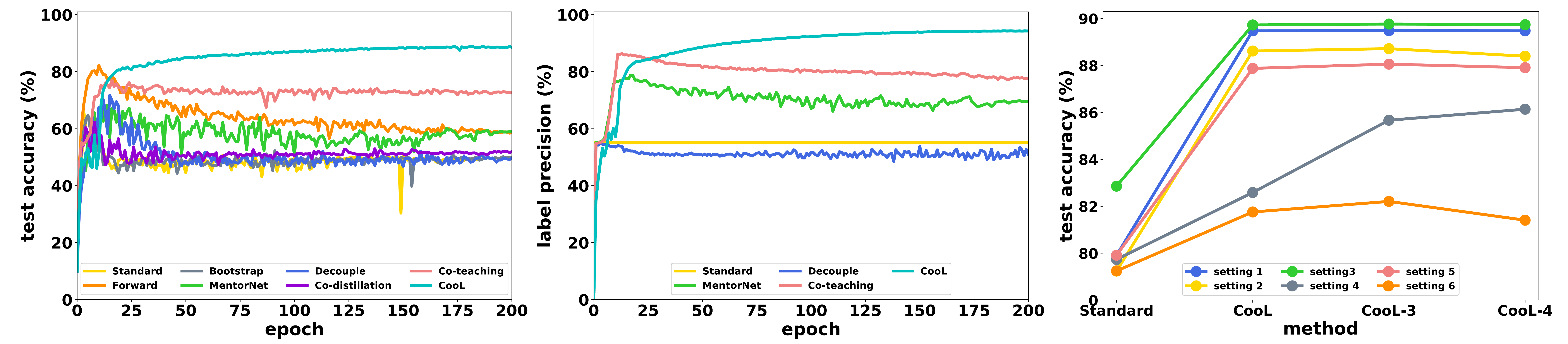}
\caption{\textbf{Left:}
Training curve on CIFAR-10 with pairwise noise ($r=0.45$);\textbf{ Middle:}
Training curve on CIFAR-10 with pairwise noise ($r=0.45$);\textbf{ Right: }Comparison of CooL, CooL-3 and CooL-4. 
}\label{fig:pn_045}
\end{figure}
Memorization effects~\cite{arpit2017closer} refer to the behavior of DNNs under noisy supervision that the model will firstly learn from clean data and eventually fit the noisy labels. This phenomenon can be visualized as rise followed by drop in the test accuracy curve.
In the left panel of Figure~\ref{fig:pn_045}, we trace the test accuracy of all baselines and CooL under high-level pairwise noise on CIFAR-10. For all baselines, the test accuracy increases at first and then decreases as the training proceeds, which matches the memorization effects. However, the curve of CooL
keeps increasing and then persists at a high level
which indicates that the proposed cooperation supervision is reliable enough to counteract the memorization effects.

\subsubsection{Reliability of the Supervision}
To further assess the reliability of the cooperation supervision 
in CooL, We report the label precision which is the ratio of correct supervisions to total supervisions. We compare CooL with the sample selection methods 
that leverage different criteria to select the 
reliable supervision.
The label precision curves in setting 2 are depicted in middle panel of Figure~\ref{fig:pn_045}. We can see that the label precision of CooL consistently increases with regard to iterations of optimizing and surpasses all the sample selection methods. In the advanced stage of training, the label precision of the cooperation supervision is 94.29\%
which means CooL can guarantee the classifiers to learn on a relatively clean dataset. This empirically verifies the reliability of the proposed cooperation supervision.

\subsubsection{Two Learners and Beyond}


Here we carry out experiments to examine our theoretical findings on the effects of utilizing multiple networks. We implement triple-network CooL (CooL-3) and quadruple-network CooL (CooL-4) and depict the accuracy along with CooL and Standard in right panel of Figure~\ref{fig:pn_045}. We can see from the curves that CooL-3 generally shows improvement or comparable results with CooL. This matches our analysis that increasing the number of the classifiers will result in a smaller risk which is closer to the lower bound. CooL-4 shows improvement in setting 4 while its accuracy slight drops in setting 2 \& 6.
This is due to insufficient information under the partition of the limited data as we have discussed formerly. As training quadruple classifiers also requires more resources, we may only resort to CooL or CooL-3 practically.

\subsection{Results on Clothing1M, Food-101N and WebVision}

\begin{table}
\begin{minipage}{0.5\columnwidth}
  \tiny
  \centering
  \setlength{\tabcolsep}{0.7mm}{
  \begin{tabular}{lrr}  

\toprule
\textbf{method} & Clothing1M & Food-101N \\
\midrule
Standard & 66.14 & 75.45\\
Forward & 67.70 & 75.92\\
Bootstrap & 66.70 & 74.75\\
MentorNet & 60.00 & 76.93\\
\midrule
Decouple & 64.44 & 72.47\\
Co-distillation & 67.23 & 77.31\\
Co-teaching & 67.30 & 77.87\\
Bagging & 67.45 & 77.14\\
\midrule
CooL & \textbf{70.79} & \textbf{80.94}\\
CooL-3 & \textbf{71.12} & \textbf{81.08}\\
\bottomrule
\end{tabular}\caption{
Results on Clothing1M and Food-101N.
}\label{tab:clothing1m}}
  \end{minipage}
\begin{minipage}{0.48\columnwidth}
  \tiny
  \centering
  \setlength{\tabcolsep}{0.9mm}{
\begin{tabular}{lrr}  
\toprule
\textbf{method} & acc.@1 & acc.@5\\
\midrule
Standard & 63.11 & 83.69\\
Forward & 63.10 & 83.78\\
LCCN & 63.52 & 84.27 \\
Bootstrap & 63.20 & 83.81\\
\midrule
Decouple & 61.23 & 81.53\\
Co-distillation & 63.41 & 84.14\\
Co-teaching & - & - \\
Bagging & 63.45 & 84.19\\
\midrule
CooL & \textbf{63.61} & \textbf{84.32}\\
CooL-3 & \textbf{63.67} & \textbf{84.39}\\
\bottomrule
\end{tabular}
\caption{
Results on WebVision.
}\label{tab:webvision}}
  \end{minipage}
\end{table}


In this section, we empirically verify the effectiveness of CooL on three large-scale datasets with real-world noise.

For Clothing1M, the results are reported in the left column of Table~\ref{tab:clothing1m}. We can see that dual-network methods generally perform better than the single versions from which they are derived. Although MentorNet does not work well in this setting, Co-teaching manages to surpass Standard by 1.16\%. Among all baselines, Forward achieves the best performance with the usage of the ground-truth transition matrix. However CooL surpasses Forward by 3.09\% without using any side information.
CooL-3 further improves CooL by 0.33\% indicating the effectiveness of leveraging multiple classifiers. 

For Food-101N, Bootstrap degenerates slightly compared to Standard, while the dual-network version method Co-distillation enjoys a 1.86\% gain. Adopting the same small loss trick, both MentorNet and Co-teaching perform well on Food-101N. MentorNet outperforms Standard by 1.48\% and Co-teaching outperforms MentorNet by 0.94\% with the use of dual-network structure. Without the knowledge of the ground-truth transition matrix, Forward only improves the test accuracy by 0.47\% compared to Standard. Again, CooL manages to outperform the best baseline by 3.07\%. Adding another classifier, CooL-3 further obtains a 0.14\% gain. 

For WebVision, we report both top-1 and top-5 accuracies in Table~\ref{tab:webvision}. The results of Co-teaching is vacant due to the absence of the ground-truth noise ratio. We can see from the results that our CooL achieves the best performance and CooL-3 further surpasses CooL. However, the gap between all the methods is trivial which may be on account of the strong open-set noise as suggested in Yao et al.~\cite{yao2019safeguarded}.

\section{Conclusion and Future Work}
In this paper, we propose a Cooperative Learning paradigm that multiple classifiers work cooperatively with noisy supervision. We demonstrate that our proposed cooperation risk is lower than that associated with individual learners. Then we present a sufficient condition where 
the risk is negatively correlated to the number of the classifiers. Finally, we introduce the Cooperative Learning framework where the reliable cooperation supervision
iteratively boosts the performance of the classifiers. We conduct a range of experiments on the CIFAR datasets
to demonstrate the robustness of CooL under synthetic noise and 
we verify the effectiveness of CooL on three real-world large-scale datasets
We further implement CooL-3 and CooL-4 to show that leveraging more classifiers can have potential gain nonetheless adding more classifiers will consume more resources. Future research directions include finding new means to generate multiple divergent classifiers to achieve lower risk and reducing the parameter space for multiple-network CooL via parameter sharing.


\small
\bibliographystyle{IEEEbib}
\bibliography{short_ref}

\end{document}